\documentclass[letterpaper, 10 pt, conference]{ieeeconf}  % Comment this line out if you need a4paper

\IEEEoverridecommandlockouts                              % This command is only needed if 
\usepackage{graphics} % for pdf, bitmapped graphics files
\usepackage{epsfig} % for postscript graphics files
\usepackage{mathptmx} % assumes new font selection scheme installed
\usepackage{times} % assumes new font selection scheme installed
\usepackage{amsmath} % assumes amsmath package installed
\usepackage{amssymb}  % assumes amsmath package installed

\usepackage{graphicx} % Required for inserting images
\usepackage{listings}
\usepackage{xcolor}
\usepackage{subcaption}
\usepackage{acronym}
\usepackage{booktabs}
\usepackage{multirow}
\usepackage{algorithm}
\usepackage{algpseudocode}
\usepackage{booktabs}

\usepackage{tikz}
\usetikzlibrary{calc,tikzmark}
\definecolor{argosgreen}{RGB}{0,255,0}

\usepackage[export]{adjustbox}

\newsavebox{\arenabox}
\newlength{\arenaht}
\newlength{\arenaw}
\newlength{\arenatop}
\newlength{\arenapad}
\newlength{\arenarowwidth}
\newcommand{\framedarenarow}[1]{%
  \sbox{\arenabox}{#1}%
  \settoheight{\arenaht}{\usebox{\arenabox}}%
  \setlength{\arenaw}{\arenarowwidth}\addtolength{\arenaw}{\arenapad}%
  \setlength{\arenatop}{\arenaht}\addtolength{\arenatop}{\arenapad}%
  \makebox[0pt][l]{%
    \begin{tikzpicture}[overlay]
      \draw[argosgreen, line width=3pt, dash pattern=on 10pt off 7pt]
            (-\arenapad,-\arenapad) rectangle (\arenaw,\arenatop);
    \end{tikzpicture}}%
  \usebox{\arenabox}%
}

\newcommand{\croppedarena}[1]{%
  \adjincludegraphics[trim={0.375\width} {0.44\height} {0.35\width} {0.30\height},
                      clip, width=\linewidth]{#1}%
}

\usepackage{hyperref}

\title{\LARGE \bf
{Auto-HSI:} Personalized human control of a robot swarm on demand\\ by using LLMs for online automatic code generation*
}

\author{A. Nazzari$^{\dagger,1,2}$, N. Cerisara$^{\dagger,1,3}$, D. Tonnis$^{\dagger,1,4}$, R. Zakir$^{1}$, L. Labarile$^{1,5}$, W. Zhu$^{1}$, \\ M. Dorigo$^{1}$, and M. K. Heinrich$^{1}$% <-this % stops a space
\thanks{*R.~Zakir, M.~Dorigo, and M.K.~Heinrich acknowledge support from the Belgian F.R.S.-FNRS, of which they are a FRIA Doctoral student, a Research Director, and a Research Associate, respectively.}% <-this % stops a space
\thanks{ $\dagger$ A. Nazzari, N. Cerisara, and D. Tonnis contributed equally to this work.}
\thanks{$^{1}$%All authors are with 
IRIDIA, Universit\'{e} Libre de Bruxelles, Brussels, Belgium %Electrical Engineering, Mathematics and Computer Science,
        {\tt\small %albert.author@papercept.net
        }}%
\thanks{$^{2}$ Department of Aerospace Science and Technology, Politecnico di Milano, Milan, Italy
        {\tt\small %email %b.d.researcher@ieee.org
        }}%
\thanks{$^{3}$ Télécom Physique Strasbourg, Université de Strasbourg, France}
\thanks{$^{4}$ Polytech Lyon, Lyon 1 Université Claude Bernard, Lyon, France }
\thanks{$^{5}$ University of Pisa, Pisa, Italy; Scuola Superiore Sant’Anna, Pisa, Italy}
}

\begin{document}

\maketitle
\thispagestyle{empty}
\pagestyle{empty}

%%%%%%%%%%%%%%%%%%%%%%%%%%%%%%%%%%%%%%%%%%%%%%%%%%%%%%%%%%%%%%%%%%%%%%%%%%%%%%%%
\begin{abstract}

This paper presents \textit{Auto-HSI}, a method for generating personalized human-swarm interaction (HSI) interfaces on demand.
The objective is to enable untrained operators to use natural language descriptions and gesture demonstrations to explain how they want the robots to collectively behave in response to their gestures. Based on these inputs, the code should automatically be generated for personalized state machines that will control the robots as desired, in response to the desired gesture inputs. In the developed \textit{Auto-HSI} prototype, the generated code produces a personalized interface for centralized control using one- and two-handed gestures, enabling a user to teleoperate the robots' motion, formation shape, and shape deformation. We test the gesture tracking and code generation components of \textit{Auto-HSI} against performance benchmarks. We then test the full \textit{Auto-HSI} prototype in ``live'' operation experiments, in which real human operators centrally control 50 simulated robots in a physics-based simulator, under nominal and noisy conditions. In these experiments, robots are teleoperated to: score a goal, traverse a maze that requires shape deformation, and score two simultaneous goals by splitting into two groups. We also demonstrate a real human operator making live updates to their personalized \textit{Auto-HSI} interface during operation (in simulation). Finally, we demonstrate live operation of real robots.

\end{abstract}

%%%%%%%%%%%%%%%%%%%%%%%%%%%%%%%%%%%%%%%%%%%%%%%%%%%%%%%%%%%%%%%%%%%%%%%%%%%%%%%%
\section{Introduction}

A key challenge in human-swarm interaction (HSI) \cite{kolling2015human} is to understand how a human operator can control a large group of robots using only one or a few inputs: that is, far fewer inputs than the total number of robots. A common approach is teleoperation via a central control station that generates
and sends commands to all or some of the robots, using inputs from gesture or pose tracking \cite{ayanian2014controlling, alonso2015gesture, macchini2021personalized, podevijn2013gesturing},
joysticks \cite{zhou2016assistive}, or haptic feedback devices \cite{lee2013semiautonomous}.
We build on this existing work and follow the same standard control station setup, focusing on teleportation using gestures.

In existing HSI studies, the set of possible swarm behaviors that an operator is able to trigger is usually predefined, and often quite limited. The interface (that is, the possible operator inputs and which robot control outputs they produce) is also usually predefined, and assumes a trained operator: the operator should already have been informed which inputs should be used to steer the swarm, and how. An important exception is~\cite{macchini2021personalized}, in which a linear model was trained to map user motion to robot motion. In this study, we aim to expand to the personalized on-demand generation of the full state machine registry and constituent code blocks that run locally on each robot, based on open-ended natural language inputs.

In our {\it Auto-HSI} approach, a personalized HSI interface is automatically generated to allow users to operate the robots in the manner they desire. In the prototype reported here, a new user needs to be instructed to use hand-based gestures, and to mark the beginning and end of a gesture with a closed fist, but the interface personalization options are otherwise open-ended. The user can request that any demonstrated gesture become associated to any of the desired multi-robot shape and motion behaviors described in natural language. 

To approach this objective, {\it Auto-HSI} makes use of existing external Large Language Models (LLMs); we demonstrate that different LLMs can be interchanged and used with our presented {\it Auto-HSI} prototype.
Automatic code generation using LLMs has been demonstrated for human control of individual robots~\cite{liang2023code}, but not yet for self-organized human-swarm interaction. Here, we demonstrate our {\it Auto-HSI} prototype for robots that can compile and run their own local copies of the generated code, but also can communicate with a central control station. Future work is to extend the \textit{Auto-HSI} prototype to the control of robot swarms that coordinate exclusively in a self-organized manner.

\section{Methods: User interaction with {\it Auto-HSI}}
\label{sec:phases}

From the user's perspective, interaction with \textit{Auto-HSI} is organized in two phases: {\it personalization}, in which the user demonstrates the desired gestures and explains in natural language the robot swarm behaviors desired for those gestures, and \emph{live operation}, in which the swarm is operated through gestures alone. The LLM is involved in the {\it personalization} phase only; it never issues commands to the robots, but rather generates and edits code that the robots execute at $10\,$Hz.
Note that the {\it personalization} phase is not restricted to system setup: the {\it personalization} phase can be called again during operation of the swarm, if the user wishes to update their personalized \textit{Auto-HSI}.

\subsection{Personalization phase}

The user demonstrates a gesture and provides a natural language description of the desired multi-robot behavior the gesture should trigger, via a generated state machine (SM) registry. The system automatically assembles a query that includes the user's description, some operating guidelines, and the current configuration of the SM registry. 
To implement the described behavior, the LLM inspects the SM registry to determine whether an existing SM already covers similar behaviors. If so, the LLM modifies the existing SM, for example by adding a node with a transition associated to the newly demonstrated gesture. Otherwise, the LLM creates a new SM. At each code generation step, the update is validated before being committed, so that syntactically invalid code never reaches the robots.

\subsection{Live operation phase}

In this phase, there are no calls to the external LLM. When a gesture is detected, it is compared via {\it dynamic time warping} (DTW) to the gesture database to determine the most similar entry: if the most similar database entry has a DTW cost below the acceptance threshold (parameter for tuning), it is considered a match.
The gesture identifier of the match is then broadcast to all robots. On the following control step each robot evaluates the outgoing transitions of its active SM node and, if a transition is associated to the broadcast identifier, it is executed. Transitions are evaluated independently in every SM, so a gesture can trigger transitions in more than one SM.

\subsection{Updates during operation: re-triggering personalization}
Once the interface is complete, the {\it personalization} phase can be re-triggered during operation if needed. The user can explain in natural language which behavior is to be modified and how; the LLM produces the corresponding change, which is applied in the same manner as in the initial {\it personalization} phase.

\section{Methods: {\it Auto-HSI} software components} \label{sec:methodology}

\begin{figure*}[htbp]
    \centering
    \includegraphics[width=1\linewidth]{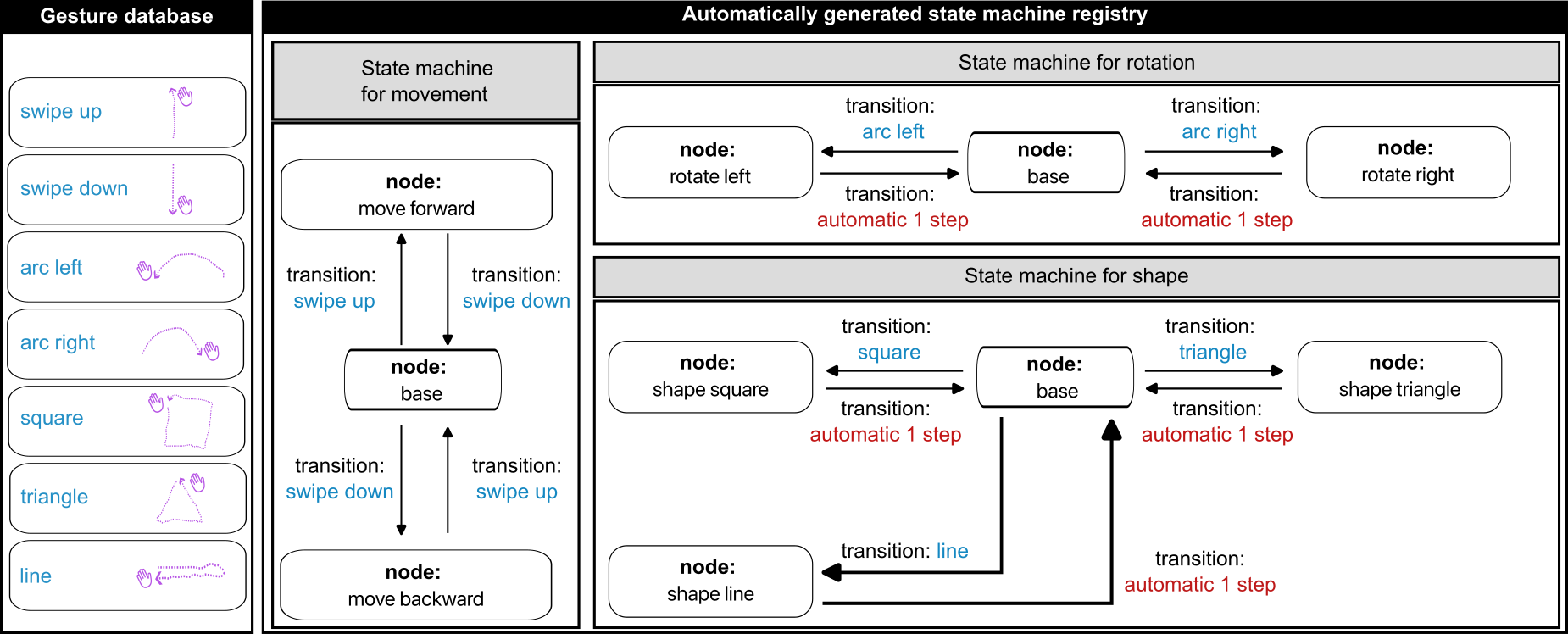}
    \caption{{\bf Diagram illustrating an example of an automatically generated {\it Auto-HSI} instance:} gesture database (left) and state machine (SM) registry (right) resulting from an example {\it personalization} phase. The gestures (labeled in blue text) demonstrated by the user are automatically associated to transitions in generated SMs for the robots' movement, formation rotation, and formation shape; these transitions are triggered by the recognition of the respective gesture. Additional transitions in the generated SMs proceed automatically (labeled in red text) after one control step and return the machine to its base node. Note that, in addition to the SM structures and transitions being automatically generated, the executable code inside the nodes of the SMs is also generated. The executable code inside the SMs can call functions to reference a database of shape definitions that is also generated, simultaneously to the generation of SMs.}
    \label{fig:gesture_sm_database}
\end{figure*}

The \textit{Auto-HSI} software includes two main internal components, gesture tracking and recognition (see Sec.~\ref{sec:gesture}) and code generation (see Sec.~\ref{sec:sm});
and two components for communicating with external elements, an external LLM (see Sec.~\ref{sec:llm_interface}) and the robots (see Sec.~\ref{sec:communication}).

The full {\it Auto-HSI} prototype, along with all software used in all experimental setups, demonstrations, and analyses of results is provided in the {\it online code repository}\footnote{Online code repository: \href{https://doi.org/10.5281/zenodo.22724336}{https://doi.org/10.5281/zenodo.22724336}}.

\subsection{Gesture tracking and recognition} \label{sec:gesture}

From each camera frame, our {\it Auto-HSI} prototype uses $21$ points $\mathbf{k}=(x_\mathbf{k},y_\mathbf{k})$ per visible hand: $4$ points (fingertip, finger base, and two intermediate joints) per finger $\mathbf{f}_i$, and $1$ for the wrist.\footnote{The 21 points-per-hand that the {\it Auto-HSI} prototype uses are extracted from camera frames using the \textit{Mediapipe} hand-landmark model~\cite{lugaresi2019mediapipe}.} 
From the 21 points $\mathbf{k} \in \mathbf{K}$ per hand, we first classify \texttt{\footnotesize fist} or \texttt{\footnotesize non-fist}, where a \texttt{\footnotesize fist} is a hand configuration $\mathbf{K}$ with no finger extended. A \texttt{\footnotesize fist} is distinguished by
comparing fingertip--wrist and finger-base--wrist distances for each finger \mbox{$\mathbf{F}\ni\{\mathbf{f}_\texttt{index},\mathbf{f}_\texttt{middle},\mathbf{f}_\texttt{ring},\mathbf{f}_\texttt{pinky}\}$}, as follows:
\begin{equation}
\small
    \mathbf{K} \in
    \begin{cases}
    \begin{aligned}
    \texttt{\footnotesize fist}~~ &\text{if~~} \forall_{\mathbf{f}_i \in \mathbf{F}}  ~~~\lVert \mathbf{k}_{\texttt{tip}_i}-\mathbf{k}_{w}\rVert \le \lVert \mathbf{k}_{\texttt{base}_i}-\mathbf{k}_{w}\rVert , \\
    \texttt{\footnotesize non-fist}~~ &\text{otherwise~~}
    \end{aligned}
    \end{cases}
\label{eq:fist}
\end{equation}
where $\mathbf{k}_{w}$ is the wrist and $\mathbf{k}_{\texttt{tip}_i}$ and $\mathbf{k}_{\texttt{base}_i}$ are the fingertip and finger-base, respectively, of finger $\mathbf{f}_i$.

For each $\textbf{K}$ classified as \texttt{\footnotesize non-fist} (Eq.~\ref{eq:fist}), we extract the center of the palm $\mathbf{c}=(x_{c},y_{c})$, defined as the geometric centroid of: $\mathbf{k}_{w}$, $\mathbf{k}_{\texttt{base}_{\texttt{index}}}$, and $\mathbf{k}_{\texttt{base}_{\texttt{pinky}}}$.
We combine the left and right palm centers, $\mathbf{c}_\texttt{left}$ and $\mathbf{c}_\texttt{right}$, into the two-handed configuration $\mathbf{C} =(x_{{c}_\texttt{left}},y_{{c}_\texttt{left}},x_{{c}_\texttt{right}},y_{{c}_\texttt{right}})$. For an undetected hand, $\mathbf{c}\leftarrow(0,0).$

Gesture detection is segmented using fists: a gesture segment is an uninterrupted sequence of frames with at least one visible $\mathbf{K} \in \texttt{\footnotesize non-fist}$, preceded and followed by frames that contain exclusively $\mathbf{K} \in \texttt{\footnotesize fist}$. After segmentation, a gesture $\mathbf{G}$ is defined as the sequence of vector differences from the segment's two-handed configurations $[\mathbf{C}_2, \mathbf{C}_n]$ to its \mbox{initial $\mathbf{C}_1$}, calculated as:
\begin{equation}
\small
    \mathbf{G} = (\mathbf{C}_{2}-\mathbf{C}_1,\; \mathbf{C}_{3}-\mathbf{C}_1,\; \dots,\; \mathbf{C}_{n}-\mathbf{C}_{1})~,
\label{eq:trajectory}
\end{equation}
such that the resulting $\mathbf{G}$ is a sequence of 4-dimensional vectors.

Gestures that are demonstrated by the user during the \textit{personalization} phase are written to a database, against which new gestures need to be compared during the \textit{live operation} phase. 
To account for the fact that the same gesture may be performed at varying speeds and thus two sequences $\mathbf{G}$ need to be aligned along the temporal axis, we score gesture similarity using DTW, calculating cumulative cost $\texttt{DTW}(i,j)$ from starting point $(1,1)$ to $(i,j)$ in the standard manner~\cite{giorgino2009computing} as:
\begin{equation}
\small
    \texttt{DTW}(i,j) = d(x_i,y_j) + \min
    \begin{cases}
        \texttt{DTW}(i-1,j) \\
        \texttt{DTW}(i,j-1) \\
        \texttt{DTW}(i-1,j-1)
    \end{cases},
\label{eq:dtw}
\end{equation}
where
$d(x_i,y_j)$ denotes Euclidean distance between 4-dimensional vectors $x_i$ and $y_j$, and sequences
$X=(x_1,x_2, ..., x_n)$ and $Y=(y_1,y_2, ..., y_n)$ are two different gestures $\mathbf{G}_1$ and $\mathbf{G}_2$ (Eq.~\ref{eq:trajectory}).

\subsection{Code generation} \label{sec:sm}

During the \textit{personalization} phase, a registry of SMs is populated incrementally as the user defines new behaviors. An SM consists of a neutral base node and a set of behavior nodes. Each node carries code blocks executed 1) on entry, 2) on every control step, and 3) on exit. Each transition is triggered either by a gesture event or automatically after a fixed number of control steps. Transitions are routed through the base node.
For each SM, a label is generated that should indicate a general behavior category to which the SM belongs, so that a later request in the same category can be applied to the same SM in an ad hoc manner.

Fig.~\ref{fig:gesture_sm_database} shows an example registry produced by the {\it personalization} phase: seven recorded gestures and three SMs, handling movement, rotation, and shape. For example, in the SM for handling movement, the \texttt{\small swipe up} gesture moves the machine into the \texttt{\small move forward} node, whose entry block sets the initial velocity to move the robots forward, control-step block maintains that velocity, and exit block reverts the velocity to zero.

The LLM receives a request in natural language and runs a sequence of tool calls that create or modify an SM registry.
Each natural language request includes the user's natural language input, the static {\it Auto-HSI} baseline prompt, and the current state of the SM registry. The full history of interactions is also sent to the LLM, to support the ability to perform incremental changes. 

The static {\it Auto-HSI} baseline prompt is created at initialization and includes the following content. 
First, it lists the functions available to interact with the SM registry: 
\vspace{1mm}
\begin{itemize}
\small
    \item retrieve SMs from the registry; 
    \item retrieve the nodes of an SM; 
    \item create or remove an SM (including its label and initial node);
    \item create, remove, or modify a node (including its transitions, and thereby the gestures that trigger transitions, and its calls to a simultaneously generated database of shapes); 
    \item modify the three code blocks of a node (\texttt{\small on entry}; \texttt{\small on exit} and \texttt{\small on step}); 
\end{itemize}
as well as the functions to add a gesture with a generated label to a gesture database, add a generated shape definition to a shape database, or modify an existing shape definition (e.g., to deform its polygon points). Note that generated shape definitions can be based on the user drawing the desired shape with a gesture or the user's natural language description of the desired shape.
It also specifies some requests for the generated SMs, e.g., the duration of a control step and %$0.1\,$s; 
the robot coordinate frame conventions.

Second, it specifies the instructions for the generated code blocks of an SM node. A block is free-form generated Lua code that makes use of some pre-existing functions, with generated arguments.
The provided list of Lua functions to be used in generated code, whether for the SMs or the shape database, includes: create a formation shape, deform a formation shape, scale a formation shape, set the formation velocity, rotate a formation, and obtain state information (formation shape, formation rotation, formation center, number of robots).

Third, it provides two examples. 
The first example is a minimal SM registry: a single SM labeled ``movement," containing the base node and a node labeled ``forward," which has an \texttt{\small on enter} code block and \texttt{\small on exit} code bock, respectively starting and then stopping the robots' motion. 
The second example is a request from a user during a {\it personalization} phase, accompanied by the ordered list of functions that should be called as a result: inspect the SM registry and verify if a new SM should be created, inspect the gesture database and verify if the requested gesture exists, register a new gesture if needed, inspect the shape database and verify if a new shape should be created, create or modify a shape as needed, create or modify an SM as needed, update the code blocks of new SM nodes as needed, and finally verify the SM has been correctly added to the registry and can be retrieved. 

Finally, it also lists some general instructions, including: guidelines on how to interact with the user by avoiding technical terms and instead using a conversational tone; suggestions on avoiding overly complex code structures when creating SM nodes; and
prohibition to ask the user for feedback on the generated code.

\subsection{Interface to an external LLM} \label{sec:llm_interface}

The natural language input from the user can be typed or dictated through a local speech-to-text engine.
The LLM returns either a text answer, which is displayed to the operator and closes the query, or a set of tool calls, which are executed against the registry before the LLM is queried again, with the results appended to the history. Requests are issued through OpenRouter's API, which enables different external LLMs to be used interchangeably by the {\it Auto-HSI} prototype. The registry manager verifies that modifications applied to the code of any node result in valid Lua code, compiles the code fragments before accepting them, and snapshots the SM registry. 
A rejected fragment is returned to the LLM and is corrected without involving the operator, while a runtime error raised on a robot is reported back to the server, which issues an automatic repair request.
Requests to the LLM run in a dedicated thread, so that neither the interface nor the communication loop is blocked while a query is resolved.

\subsection{Communication with the robots} \label{sec:communication}

In the simulation setup, each robot holds its own TCP connection to the server and runs its own local copy of the generated code for robot control (i.e., the SM registry with its required shapes).
When a modification of one SM is committed, the affected SM is serialized to JSON and pushed over the open sockets. A full node update rebuilds the SM in memory while preserving the identifier of the active node. A partial node update recompiles only the relevant code blocks.

\section{Results}

First, the two main internal components of {\it Auto-HSI} (gesture tracking and recognition; code generation), are evaluated against performance benchmarks. Then, we present experiments and demonstrations of real human operators\footnote{Note that, in this paper, black boxes have been added to images containing faces, for de-identification.} using the full {\it Auto-HSI} prototype. 

All experimental results data, raw videos, and LLM logs are provided in the {\it online data repository}\footnote{Online data repository: \href{https://doi.org/10.5281/zenodo.22727703}{https://doi.org/10.5281/zenodo.22727703}. \textbf{[Note for double-blind review: Temporarily, repository has been partially redacted because of faces of authors in raw experiment videos.]}}.

\subsection{Benchmark: Gesture recognition}

To assess the gesture recognition performance of the tracking and recognition method (Sec.~\ref{sec:gesture}), we use a benchmark matrix $\mathcal{G}_{m\times n}^*$ 
that consists of \mbox{$n=41$} ground-truth gestures (for descriptions of the 41 ground-truth gestures, see Table~S1 in the {\it online data repository})) performed $m=4$ times, by 4 different individual operators, for a total of 164 ground-truth recordings $\mathbf{G}_{ij}^{*} \in \mathcal{G}^*$. 
The test matrix $\widehat{\mathcal{G}}_{m\times n}$ 
consists of the same \mbox{$n=41$} gestures, but performed $m=16$ times (4 times each by 4 operators), for a total of 656 test recordings $\widehat{\mathbf{G}}_{ij} \in \widehat{\mathcal{G}}$. 

For each test sample $\widehat{\mathbf{G}}_{ij} \in \widehat{\mathcal{G}}$, we match it to the ground-truth sample $\mathbf{G}_{ij}^{*} \in \mathcal{G}^*$ that has the lowest \texttt{DTW} cost according to Eq.~\ref{eq:dtw}. We then assess performance according to: true positives (\texttt{TP}), when the test column $\widehat{j}$ (i.e., $\widehat{\mathbf{G}}_{:j} \in \widehat{\mathcal{G}}$) was judged to match the selected ground-truth column $j^*$ (i.e., $\mathbf{G}_{:j}^{*} \in \mathcal{G}^*$) and the judgment is correct; false positives (\texttt{FP}), when $\widehat{j}$ was judged to match $j^*$ but does not; and false negatives (\texttt{FN}), when $\widehat{j}$ was judged to not match $j^*$ but does.

We assess performance according to recall $R_{\widehat{j}}$~, the true positive rate for test column $\widehat{j}$, calculated as:
\begin{equation}
\small
     R_{\widehat{j}} = \texttt{TP}_{\widehat{j}} ~/~ (\texttt{TP}_{\widehat{j}} + \texttt{FN}_{\widehat{j}})~,
\label{eq:recall}
\end{equation}
and precision $P_{j^*}$ for ground-truth column $j^*$, calculated as:
\begin{equation}
\small
    P_{j^*} = \texttt{TP}_{j^*} ~/~ (\texttt{TP}_{j^*} + \texttt{FP}_{j^*})~.
\label{eq:precision}
\end{equation}
Recall $R$ measures the probability that the personalized {\it Auto-HSI} interface correctly identifies a gesture performed during the {\it live operation} phase against those demonstrated during the {\it personalization} phase. %in other words, the probability that the operator is able to trigger the robot behavior intended.
Low recall would result in robots failing to perform a behavior requested by the operator.
Precision $P$ measures the probability that a gesture selected from among those demonstrated during the {\it personalization} phase is indeed the gesture performed during the {\it live operation} phase. %the operator had actually requested it
Low precision would result in robots performing a behavior that was not requested.

Table~\ref{tab:results_gesture_benchmark} reports the results, with gestures (i.e., ground-truth column $j^*$ and its accompanying test column $\widehat{j}$) categorized into four gesture types: one-handed swipes, two-handed swipes, one-handed shapes, and two-handed shapes. 

\begin{table}[htbp]
\centering
\caption{Gesture recognition performance by gesture type, with the number of ground-truth columns $j^*$ (``Qty") and test samples $\widehat{i,j}$ (``Trials") listed for each gesture type. We report recall $R$ (Eq.~\ref{eq:recall}), 
and precision $P$ (Eq.~\ref{eq:precision}), with their 95\% confidence intervals. Recognition is tested against the full confusion matrix, i.e.\ each of the 656 test samples is matched against all 164 ground-truth samples, no matter the gesture type.}
\label{tab:results_gesture_benchmark}
\setlength{\tabcolsep}{2pt}
\begin{tabular}{@{}l|c|c|c c|c c@{}}
\toprule
{\bf Gesture type} & {\bf Qty} & {\bf Trials} &  $R$,\,Eq.\,\ref{eq:recall}, & 95\%\,CI (\%) & $P$,\,Eq.\,\ref{eq:precision}, & 95\%\,CI (\%) \\
\midrule
One-hand swipe~ & 16 & 256 & 99.2 & [97.2, 99.8] & 97.7& [95.1, 98.9] \\
Two-hand swipe~ & 10 & 160 & 95.0 & [90.4, 97.4] & 93.3 & [88.3, 96.2]\\
One-hand shape~ & 8 & 128 & 90.6 & [84.3, 94.6] & 90.6 & [84.3, 94.6]\\
Two-hand shape~ & 7 & 112 & 86.6 & [79.1, 91.7] & 92.4 & [85.7, 96.1] \\
\midrule
{\bf Overall} & 41 & 656 & 94.4 & [92.3, 95.9] & 94.4 & [92.3, 95.9]\\
\bottomrule
\end{tabular}
\end{table}

The results in Table~\ref{tab:results_gesture_benchmark} show that swipe gestures are recognized reliably, both one- and two-handed, with one-handed swipes reaching a recall of $99.2\%$ (254 of 256 test samples). As expected, the more complex shape gestures, both one- and two-handed, are recognized less reliably, although performance remains high: recall drops to $90.6\%$ and $86.6\%$ respectively, the latter being the lowest value observed across gesture types. Overall, 619 of 656 test samples are classified correctly, corresponding to an accuracy of $94.4\%$ ($95\%$ CI $[92.3, 95.9]$).
%For all gesture types, precision closely follows recall, indicating that no gesture type systematically attracts misclassifications from the other types.

\subsection{Benchmark: Code generation}

To assess the code generation method (Sec.~\ref{sec:sm}), we test the reliability of the robot behaviors that result from the generated SMs (including the generated SM structure and its generated code blocks). 

The benchmark is structured in four baseline categories: basic motion, shape deformation, shape creation, and SM modification. Each category includes
2--5 baseline SMs that were manually designed to fulfill a pre-defined behavior description, for a total of 15 baseline SMs (for descriptions of the 15 baseline SMs, see Table~S3 in the {\it online data repository}). 
For each baseline SM, a set of test SMs is generated from a natural language description of the desired behavior.
We run 18 test trials per baseline SM (a total of 270 trials), with the following variations.
For each baseline SM, we test three different amounts of detail in the natural language prompt (high, mid, or low), with two different prompt versions (single-message and multi-message), for a total of 6 test prompts per baseline SM. Each test prompt is input to three different example external LLMs: DeepSeek v4 Pro, DeepSeek v4 Flash, and Gemma 4 31B. 

\begin{table}[htbp]
\centering
\caption{Percentage of the code generation trials, per experimental variation, in which the robot behaviors produced by the test SMs are sufficiently visually similar to those produced by the baseline SMs, according to the test metrics (for threshold details, see Table~S2 in the {\it online data repository}). 
The three example external LLMs are DeepSeek v4 Pro (LLM-1), DeepSeek v4 Flash (LLM-2) and Gemma 4 31B (LLM-3). A \mbox{checkmark ($\checkmark$)} indicates 100\% of trials in that variation. There are 270 trials in total.}
\label{tab:results_llm_benchmark}
\setlength{\tabcolsep}{4pt}
\begin{tabular}{cccccc}
\toprule
{\bf Category} & {\bf Metric} & {\bf Prompt detail} & {\bf LLM-1} & {\bf LLM-2} & {\bf LLM-3} \\
\midrule
\multirow{3}{*}{\shortstack[c]{{\bf Motion}\\ \textit{(72 trials)}}} & \multirow{3}{*}{$d_\texttt{track}$} & High & $\checkmark$ & $\checkmark$ & $\checkmark$ \\
                     && Mid  & $\checkmark$ & $\checkmark$ & $\checkmark$ \\
                     && Low  & $\checkmark$ & $\checkmark$ & $\checkmark$ \\
\midrule
\multirow{3}{*}{\shortstack[c]{{\bf Deform}\\{\bf shape}\\ \textit{(90 trials)}}} & \multirow{3}{*}{$d_\texttt{shape}$} & High & $\checkmark$ & $\checkmark$ & $\checkmark$ \\
                     && Mid  & $\checkmark$ & $\checkmark$ & 80\% \\
                     && Low  & $90\%$ & $90\%$ & $77.8\%$ \\
\midrule
\multirow{3}{*}{\shortstack[c]{{\bf Create}\\{\bf shape}\\ \textit{(72 trials)}}} & \multirow{3}{*}{$d_\texttt{shape}^\lambda$} & High & $\checkmark$ & 75\% & $\checkmark$ \\
                     && Mid  & $62.5\%$ & $50\%$ & $87.5\%$ \\
                     && Low  & $75\%$ & $62.5\%$ & $75\%$ \\
\midrule
\multirow{3}{*}{\shortstack[c]{{\bf Modify SM}\\ \textit{(36 trials)}}} & \multirow{3}{*}{$d_\texttt{shape}$} & High & $\checkmark$ & $\checkmark$ & $\checkmark$ \\
                     && Mid  & $\checkmark$ & $\checkmark$ & $\checkmark$ \\
                     && Low  & $\checkmark$ & $\checkmark$ & $\checkmark$ \\
\bottomrule
\end{tabular}
\end{table}

We compare the output of the generated and the baseline SM in numerical simulation, by comparing the target robot positions produced by the SMs, in terms of both shape and movement over time. 
We make the comparisons using two metrics. First, motion tracking error $d_\texttt{track}$, which we define as the Euclidean distance between the geometric centroids of the robots' target positions, at each time step.
Second, the formation error $d_\texttt{shape}$,
which compares the inter-robot distances of the robots' target positions. 
To calculate $d_\texttt{shape}$, we measure the Euclidean distance between each robot position $i$ and every other robot position $j$ (when the distances are normalized, we denote the metric as $d_\texttt{shape}^\lambda$). For each formation, we then sort the list of distances in ascending order.
The formation error $d_\texttt{shape}$ then is the sum of absolute differences between the elements of the two lists.

The multi-robot behaviors produced by the test SMs are sufficiently visually similar to those produced by the baseline SMs , and thus are considered successful generations of the desired multi-robot behavior, in 91.4\% of the 270 trials. As seen in Table~\ref{tab:results_llm_benchmark}, shape creation is the most demanding category: with medium and low levels of prompt detail, the desired robot behaviors are produced in only 68.75\% of trials. This is likely attributable to shape creation involving far more degrees of freedom, compared to defining motion directions or modifying an existing shape or SM. 

The fact that the desired behaviors are not always produced in all trials, even when using the highest level of prompt detail, necessitates our {\it Auto-HSI} design choice to conduct the {\it personalization} phase before the {\it live operation} phase, and to include a multi-robot simulator in-the-loop in the {\it personalization} phase, to provide feedback to the operator and enable multi-shot queries until the produced robot behaviors match what is desired.

\subsection{Real human teleoperation of 50 simulated robots}

We test teleoperation of simulated robots by real human operators, using the full {\it Auto-HSI} prototype, in three different tasks: scoring a goal (Task-1), traversing a maze with narrow gaps (Task-2), and scoring two goals at the same time (Task-3). Together, the three tasks demonstrate personalized teleoperation of the robot swarm's formation shape, motion trajectory, shape deformation, and splitting into two groups. We use 50 differential-drive ground robots, modeled after the \mbox{e-puck~\cite{mondada2009puck}} robot, in the physics-based simulator ARGoS~\cite{pinciroli2012argos}.

Each task is tested: a) in the nominal condition, i.e., no noise injection; and b) with noisy motion, added using the noise parameter $e = 1.5$. At each time step, noise is added independently to each wheel speed $w$ of each robot (i.e., 100 wheels in total), sampled uniformly randomly, as follows: 
\begin{equation}
\small
    w_i \leftarrow w_i + |w_i| \cdot \texttt{\small Uniform}[{-e, e}]~.
\end{equation}

Each task and noise condition has been executed by three different operators (i.e., a total of 18 trials). The operators each approach the {\it personalization} phase differently, demonstrating different gestures and explaining different desired meanings in natural language, thus resulting in three distinct personalized {\it Auto-HSI} interfaces. The operators use their different personalized interfaces to complete the same tasks, but in different manners (see an example of this personalization in Fig.~\ref{fig:griglia}).

\subsubsection{Scoring a goal \textbf{(Task-1)}}

\begin{figure*}[htbp]
    \centering
    \begin{subfigure}{0.48\textwidth}
        \includegraphics[height=0.5\textwidth]{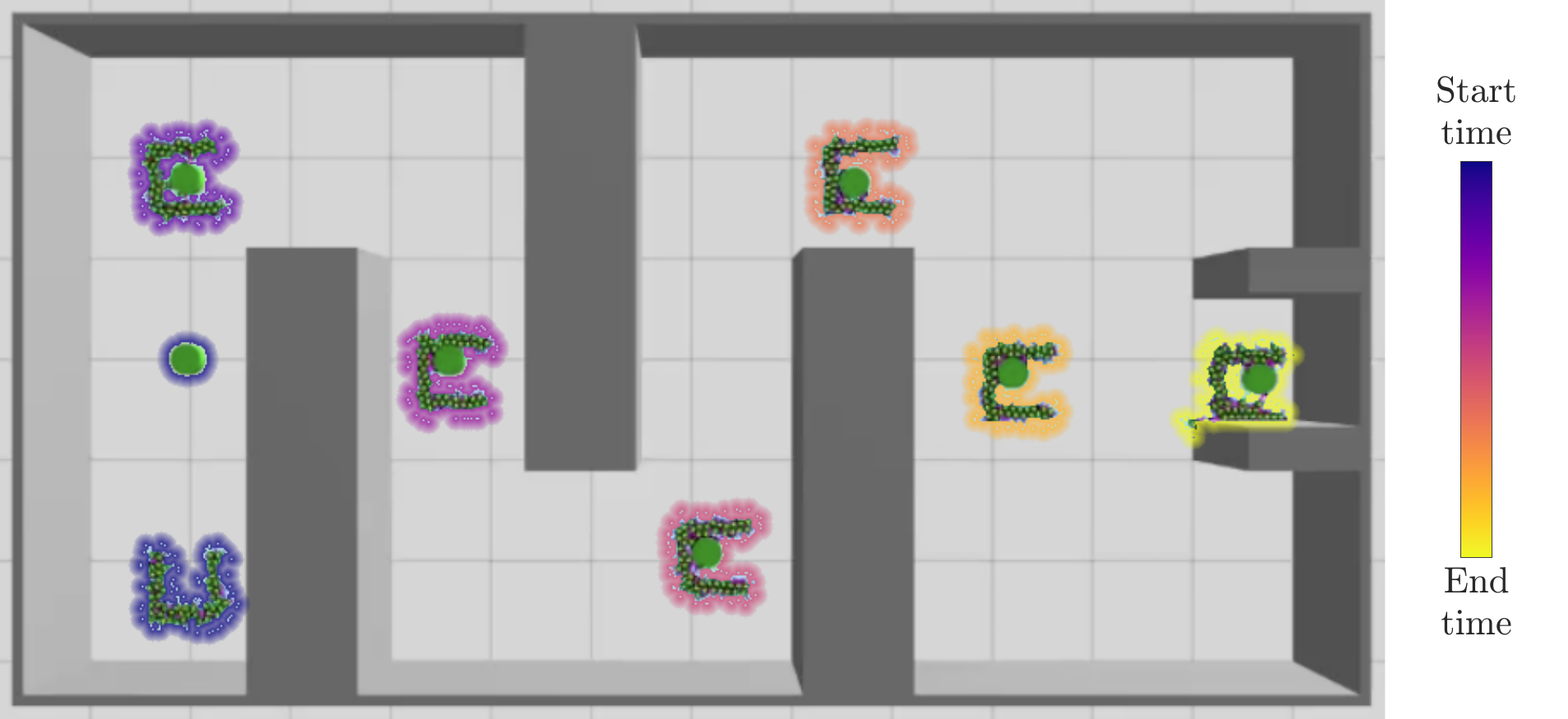}
        \caption{Nominal condition; no noise ($e = 0.0$), 30\,s frame interval.~~~~}
        \label{fig:demo1_noerror}
    \end{subfigure}\hfill
    \begin{subfigure}{0.48\textwidth}
        \includegraphics[height=0.5\textwidth]{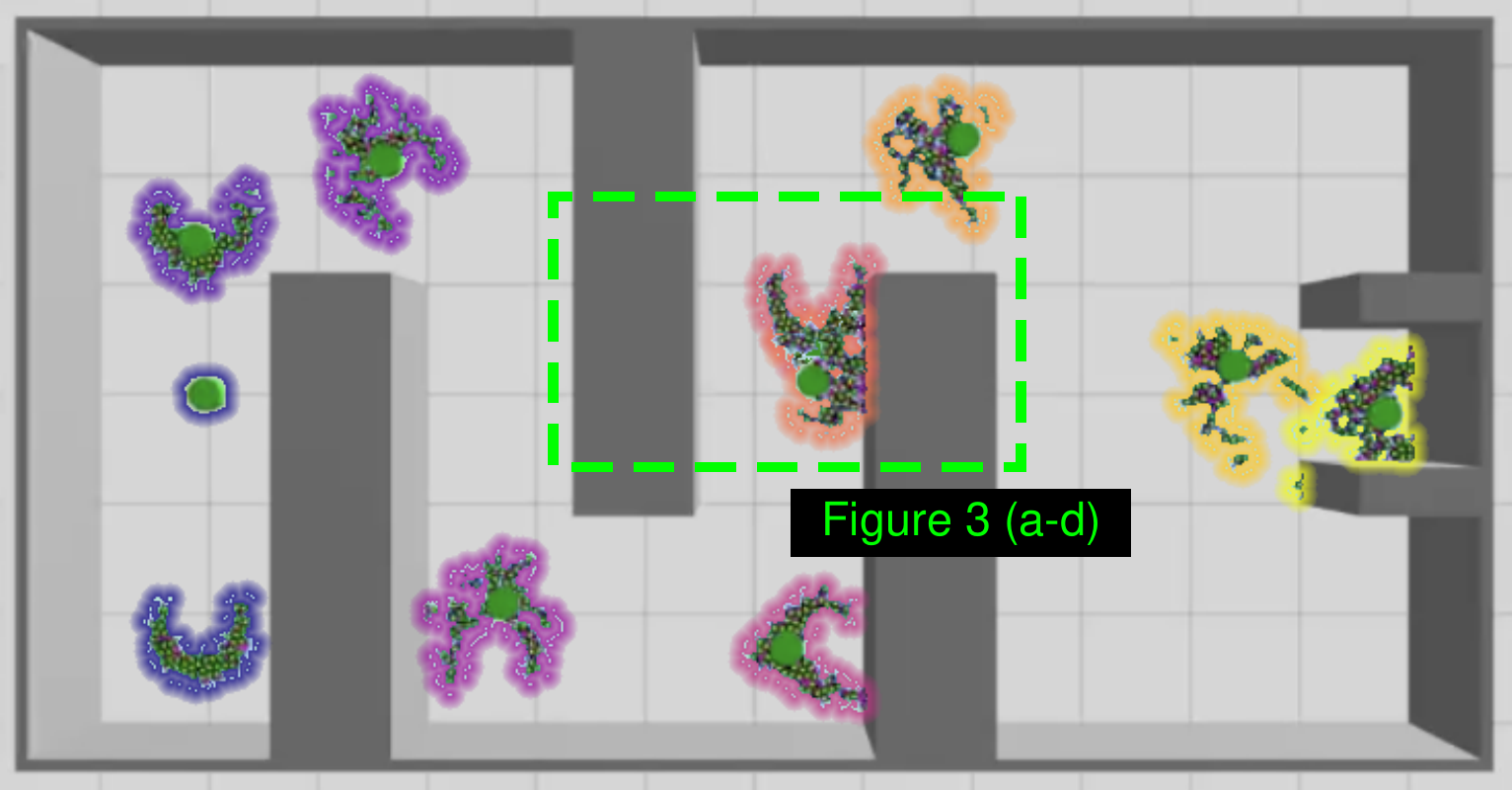}
        \caption{Condition with wheel-speed noise ($e = 1.5$), 30\,s frame interval.}
        \label{fig:demo1_error}
    \end{subfigure}
    \caption{{\bf Scoring a goal (Task-1), video timelapse:} example trials of the same operator in both the (a) nominal condition and (b) noisy condition. The task is to use the robots to push a ball (dark-green circle) through an environment with obstacles, then into the goal position at the end (i.e., the small gap between two parallel walls at the right-hand boundary of the environment). (b) In the condition with wheel-speed noise, it is more difficult for the operator to retain control of the ball. At one point (see light-green dashed box), the operator loses the ball due to the wheel-speed noise and is forced to rotate the formation to recapture it. Additional time steps of this maneuver are shown in Fig.~\ref{fig:states_gestures}.}
    \label{fig:demo_1_full_image}
\end{figure*}

\begin{figure*}[htbp]
    \centering
    \newlength{\arenaimg}\setlength{\arenaimg}{0.19\textwidth}%
    \newlength{\arenagap}\setlength{\arenagap}{0.0125\textwidth}%
    \setlength{\arenarowwidth}{0.7975\textwidth}

    \begin{subfigure}[t]{\arenaimg}
        \framedarenarow{\croppedarena{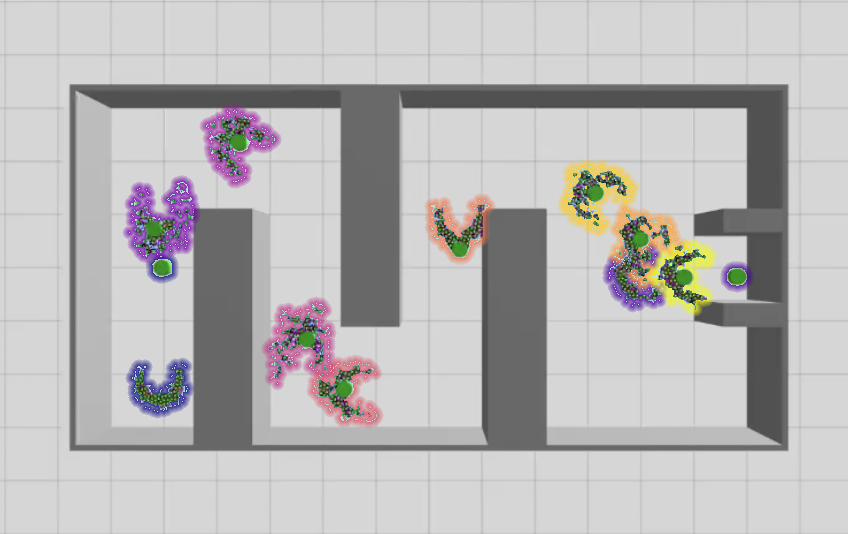}}
        \caption{Move forward {\bf (155\,s)}}
        \label{fig:arena2}
    \end{subfigure}\hspace{\arenagap}%
    \begin{subfigure}[t]{\arenaimg}
        \croppedarena{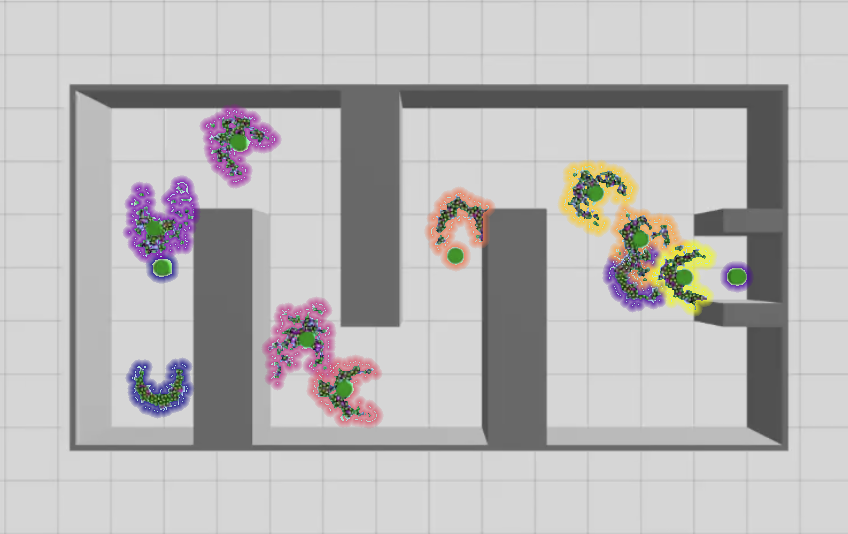}
        \caption{Rotate {\bf (167\,s)}}
        \label{fig:arena3}
    \end{subfigure}\hspace{\arenagap}%
    \begin{subfigure}[t]{\arenaimg}
        \croppedarena{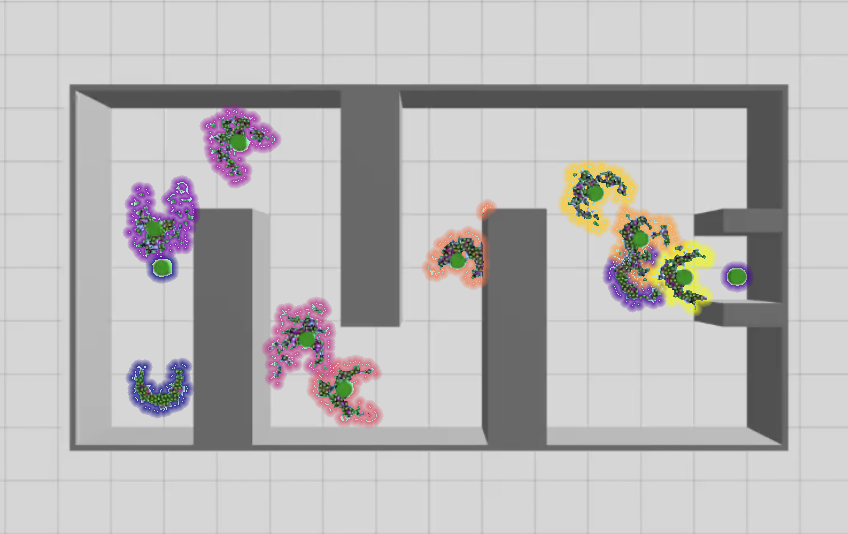}
        \caption{Move forward {\bf (176\,s)}}
        \label{fig:arena4}
    \end{subfigure}\hspace{\arenagap}%
    \begin{subfigure}[t]{\arenaimg}
        \croppedarena{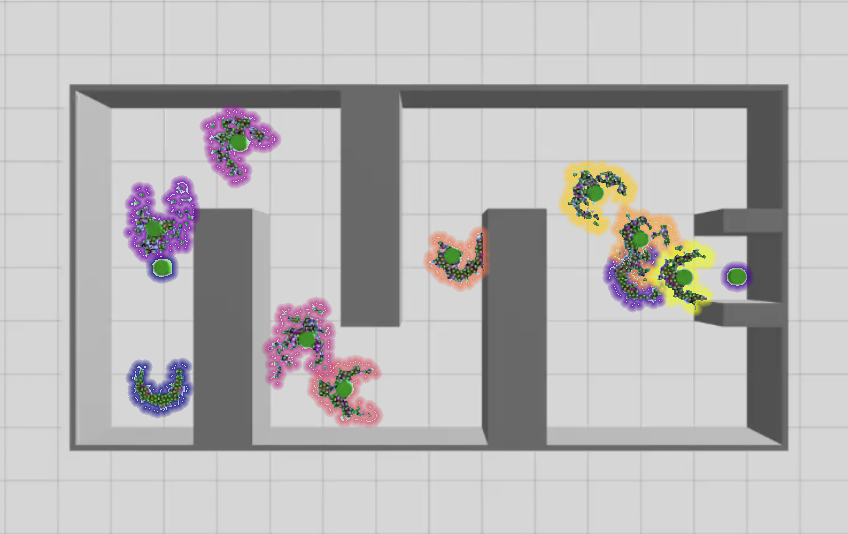}
        \caption{Rotate {\bf (189\,s)}}
        \label{fig:arena5}
    \end{subfigure}\\[0.3em]%
    \begin{subfigure}[t]{\arenaimg}
        \includegraphics[width=\linewidth]{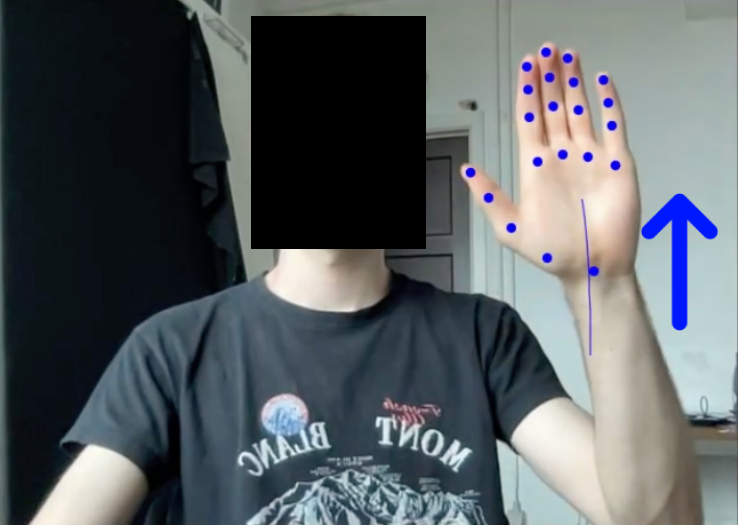}
        \caption{Swipe up {\bf (130\,s)}}
        \label{fig:pilota1}
    \end{subfigure}\hspace{\arenagap}%
    \begin{subfigure}[t]{\arenaimg}
        \includegraphics[width=\linewidth]{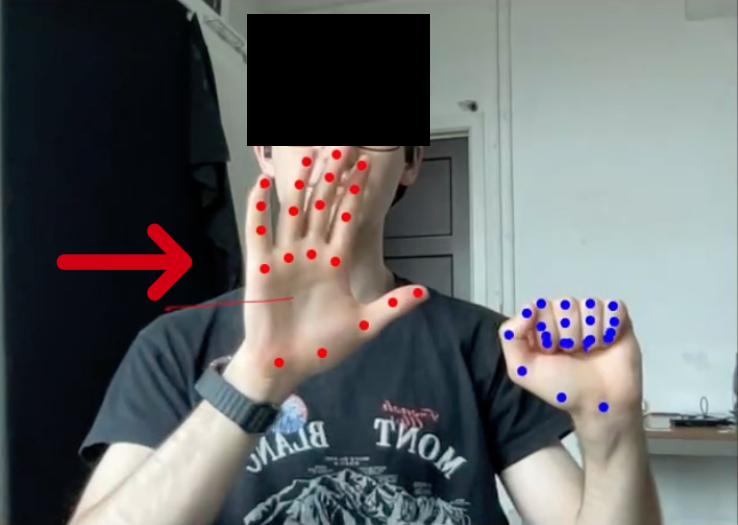}
        \caption{Swipe right {\bf (157\,s)}}
        \label{fig:pilota2}
    \end{subfigure}\hspace{\arenagap}%
    \begin{subfigure}[t]{\arenaimg}
        \includegraphics[width=\linewidth]{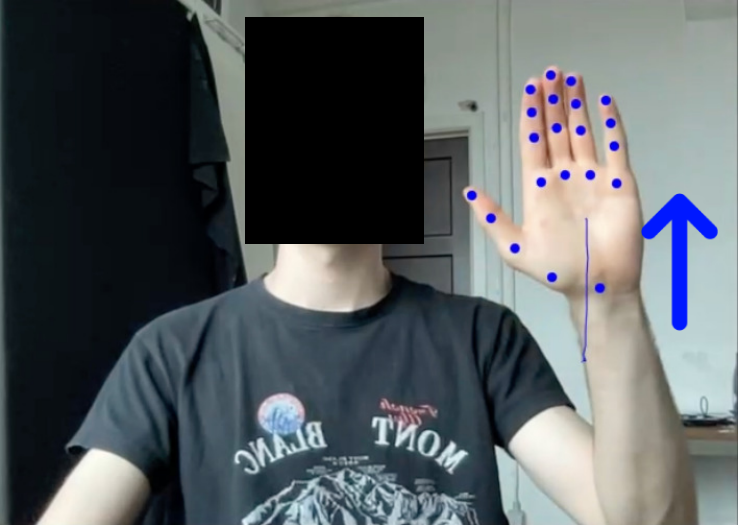}
        \caption{Swipe up {\bf (168\,s)}}
        \label{fig:pilota3}
    \end{subfigure}\hspace{\arenagap}%
    \begin{subfigure}[t]{\arenaimg}
        \includegraphics[width=\linewidth]{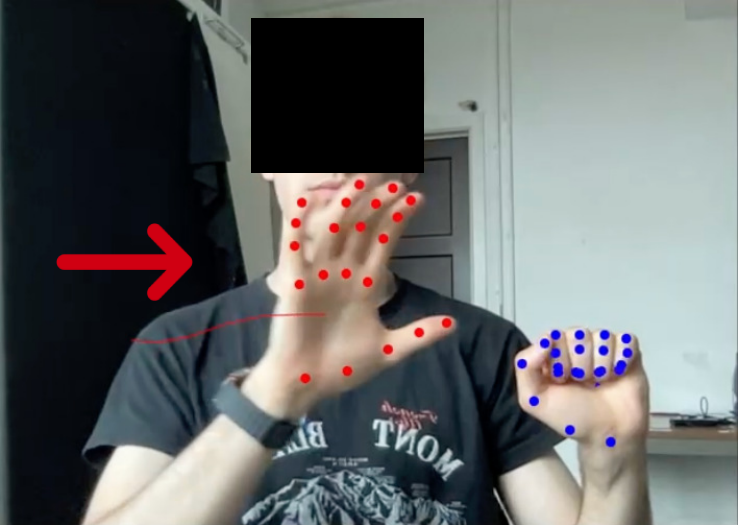}
        \caption{Swipe right {\bf (178\,s)}}
        \label{fig:pilota4}
    \end{subfigure}
    \caption{{\bf Highlight from Task-1 (Fig.~\ref{fig:demo1_error}, time 130--178\,s): operator performs ball recovery to compensate for robots' wheel-speed noise.} (a--d) Zoom-in views of the area in the light-green dashed box in Fig.~\ref{fig:demo1_error}, with (e--h) operator gestures that triggered their motion. 
    (a) Ball slips through a gap between robots while they move forward in response to (e) {\it swipe up}. Operator needs to turn back to retrieve the ball. Robots (b) rotate in response to (f) {\it swipe right}, then (c) move forward in response to (g) {\it swipe up}. When the operator judges the ball to be enclosed again, the task needs to be resumed: robots (d) rotate in response to (h) {\it swipe right}. Note that gestures are described according to camera view, not operator view.
    }
    \label{fig:states_gestures}
\end{figure*}

\begin{figure*}[htbp]
    \centering
    \begin{subfigure}{0.48\textwidth}
        \includegraphics[height=0.5\textwidth]{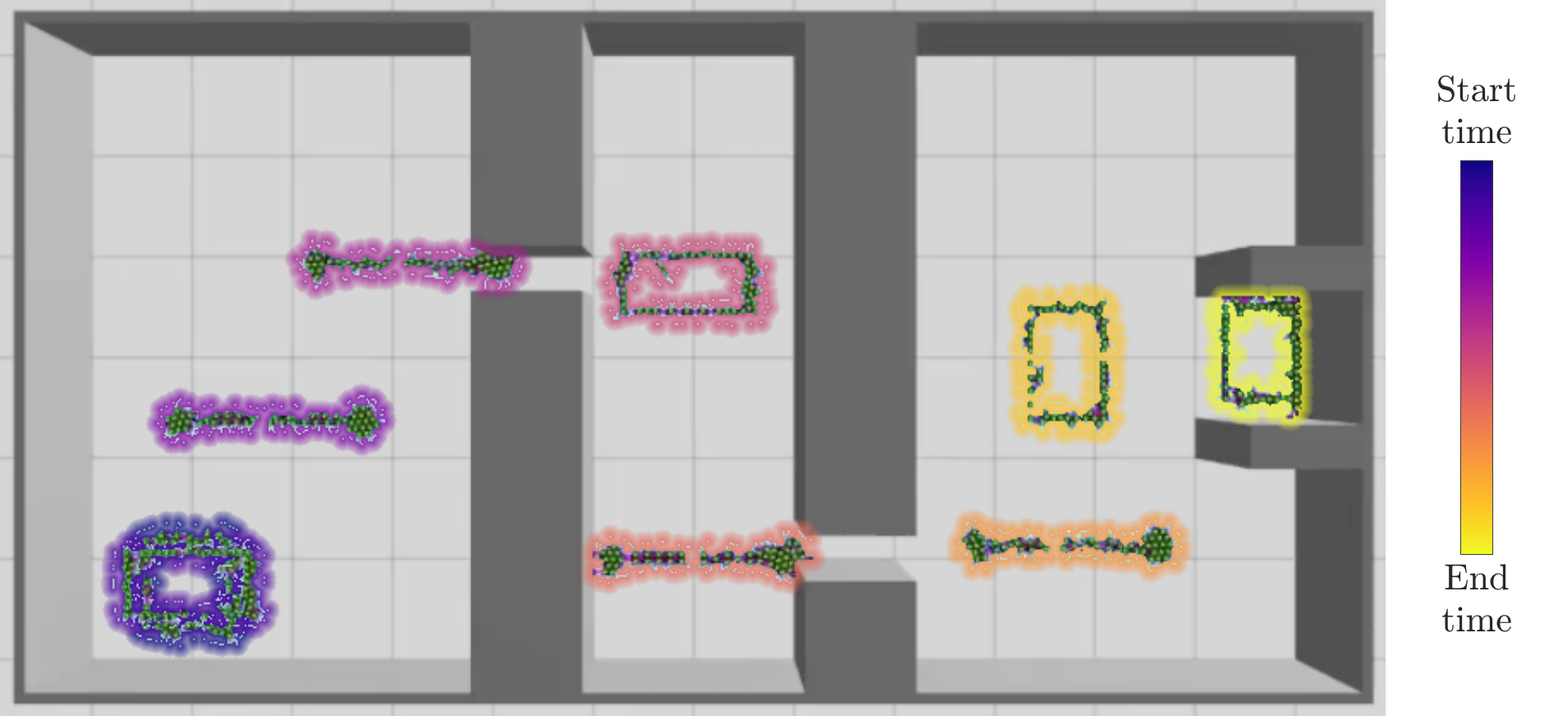}
        \caption{Task-2, 
        30\,s frame interval.}
        \label{fig:demo2_arena}
    \end{subfigure}\hfill
    \begin{subfigure}{0.48\textwidth}
        \includegraphics[height=0.5\textwidth]{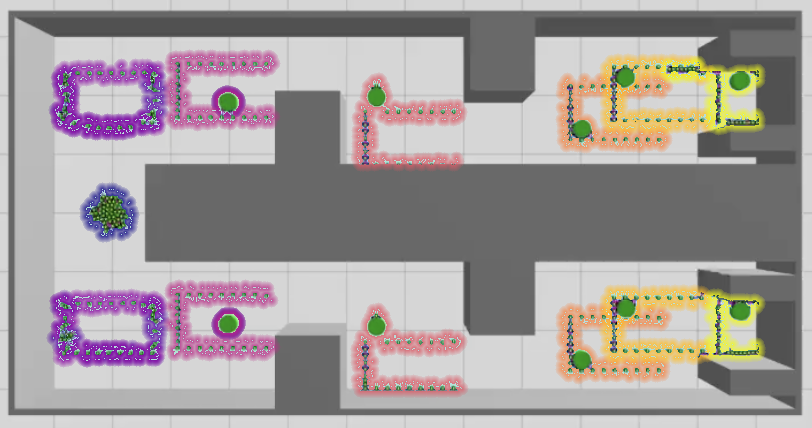}
        \caption{Task-3, 
        30\,s frame interval.}
        \label{fig:demo3_arena}
    \end{subfigure}
    \caption{\textbf{Traversing a maze with narrow gaps (Task-2) and scoring two goals at the same time (Task-3), video timelapse:} example trials in the nominal condition (no noise; $e=0.0$). (a) In {\bf Task-2}, the task is to start with a large shape and then deform it to fit through narrow gaps in the environment, until reaching the goal position at the end (at the right-hand boundary). (b) In {\bf Task-3}, the task is to split the robots into two groups, then use them to push two different balls (dark-green circles) along two identical paths simultaneously, then into two different goal positions (at the right-hand boundary).}
    \label{fig:demos2-3}
\end{figure*}

In Task-1, the user needs to navigate through an arena to move a ball to a target location (example trials shown in Fig.~\ref{fig:demo_1_full_image}).

\subsubsection{Traversing a maze with narrow gaps \textbf{(Task-2)}}

In Task-2, the user needs to deform the robots' formation shape to traverse a maze with two narrow gaps and reach a target location (example trial shown in Fig.~\ref{fig:demo2_arena}).
The operator initially forms the desired rectangular shape, then compresses it vertically when passing through a narrow gap, then compresses it horizontally to navigate between the walls to get to the next narrow gap. To navigate through the different areas of the environment, the operator must iteratively deform the shape in a trial-and-error process to get it to fit and not lose robots from getting stuck along the way (see full videos in the {\it online data repository}). 

\subsubsection{Scoring two goals at the same time \textbf{(Task-3)}}

In Task-3, the user needs to split the robots into two groups and navigate both groups through an arena to move two balls into two target locations (example trial shown in Fig.~\ref{fig:demo3_arena}).

\subsubsection{Results of all trials \textbf{(Tasks-1, -2, and -3})}

In all 18 trials (i.e., both with and without wheel-speed noise), the operators are able to complete the task.

To summarize the comparison between the conditions with and without wheel-speed noise, Table~\ref{tab:demo_comparison} reports the completion time and number of utilized gestures for each of the three operators, for each noise condition. The completion time is measured from the execution of the first gesture.
 
In the nominal condition ($e = 0$) the three operators perform comparably: completion times range from 123\,s to 241\,s and the number of gestures from 22 to 54, with no operator notably outperforming the other two. 

In the condition with wheel-speed noise ($e = 1.5$), performance compared to the nominal condition is similar (or sometimes a little better) in Task-2, while notably worse in Task-1 and Task-3. This can be attributed to the fact that corrective actions in Task-2 manage the robots only, while corrective actions in Tasks-2 and -3 also need to manage the ball(s) that might need to be retrieved (see Fig.~\ref{fig:states_gestures}) The performance degradation varies notably between operators in Task-1: User-1 requires $1.5\times$ the time and $2.1\times$ the gestures used in the nominal condition, compared to $2.0\times$ and $3.7\times$ for User-3 and $2.3\times$ and $4.6\times$ for User-2. 
In Task-3, User-1 and User-2 require $1.8\times$ to $1.9\times$ the time as in the nominal condition. User-3 shows the worst degradation, requiring 720\,s and 205 gestures, compared to 125\,s and 26 gestures in the nominal condition. 

\begin{table}[htbp]
    \centering
    \caption{Comparison of completion time and number of gestures used in the three tasks, for three different human operators (User-1, User-2, User-3).}
    \label{tab:demo_comparison}
    \setlength{\tabcolsep}{3pt}
    \begin{tabular}{c c ccc ccc}
        \toprule
        & & \multicolumn{3}{c}{\textbf{Completion time (s)}} & \multicolumn{3}{c}{\textbf{Number of gestures}} \\
        \cmidrule(lr){3-5} \cmidrule(lr){6-8}
        \textbf{Task} & {\bf Noise}\,$e$~ & User-1 & User-2 & User-3 & User-1 & User-2 & User-3 \\
        \midrule
        \multirow{2}{*}{Task-1} & 0.0 & 185 & 241 & 157 & 22 & 36  & 23 \\
                           & 1.5 & 270 & 549 & 308 & 46 & 164 & 84 \\
        \midrule
        \multirow{2}{*}{Task-2} & 0.0 & 237 & 205 & 123 & 49 & 44 & 23 \\
                           & 1.5 & 197 & 186 & 153 & 44 & 46 & 27 \\
        \midrule
        \multirow{2}{*}{Task-3} & 0.0 & 150 & 210 & 125 & 26 & 54 & 26 \\
                           & 1.5 & 270 & 400 & 720 & 71 & 93 & 205 \\
        \bottomrule
    \end{tabular}
\end{table}

\begin{figure}[t]
    \centering
    \begin{subfigure}[b]{0.45\columnwidth}
        \centering
        \includegraphics[trim=0 50 0 55, clip, width=\linewidth]{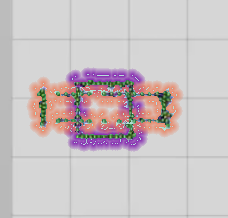}
        \caption{{\bf Trial 1} of shape deformation (from purple to orange)}
        \label{fig:img1_samegestures}
    \end{subfigure}
    \hspace{1mm}
    \begin{subfigure}[b]{0.45\columnwidth}
        \centering
        \includegraphics[trim=0 60 0 45, clip, width=\linewidth]{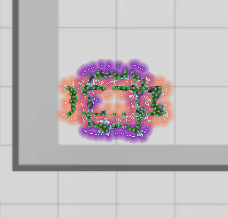}
        \caption{{\bf Trial 2} of shape deformation (from purple to orange)}
        \label{fig:img3}
    \end{subfigure}

    \vspace{0.5em}

    \begin{subfigure}[b]{0.45\columnwidth}
        \centering
        \includegraphics[trim=0 10 0 55, clip, width=\linewidth]{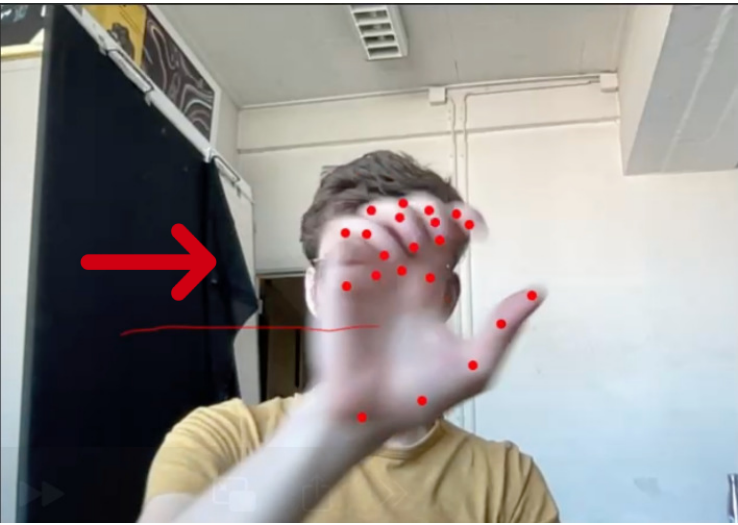}
        \caption{Swipe right}
        \label{fig:img2}
    \end{subfigure}
    \hspace{1mm}    
    \begin{subfigure}[b]{0.45\columnwidth}
        \centering
        \includegraphics[trim=0 10 0 55, clip, width=\linewidth]{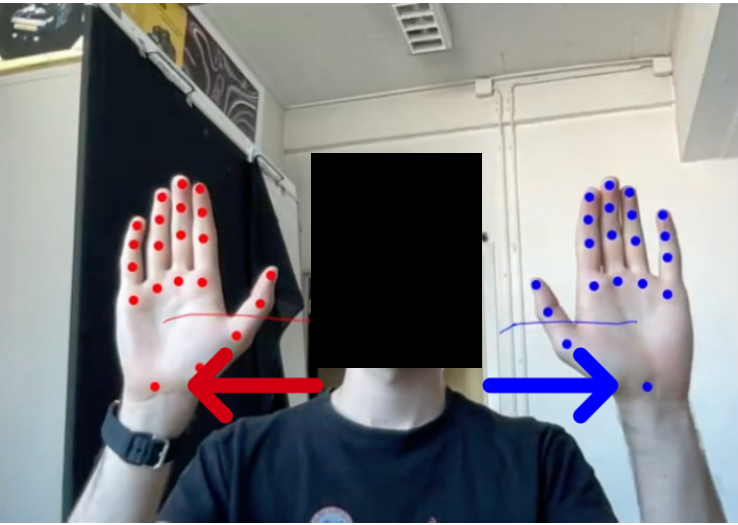}
        \caption{Swipe outwards}
        \label{fig:img4}
    \end{subfigure}

    \caption{ {\bf Highlight from Task-2: two different operators' personalized {\it Auto-HSI} interfaces.} In both example trials, (a--b) the robots' shape is being deformed in a similar manner, from approximately a square (purple) to a more elongated rectangle (orange), although two different gestures are used: (a) is produced by gesture (c) {\it swipe right}, and (b) is produced by gesture (d) {\it swipe outwards}. Note that gestures are described according to camera view, not operator view.}
    \label{fig:griglia}
\end{figure}

\subsection{Demo: Re-triggering personalization during operation}

We perform a demonstration of re-triggering the {\it personalization} phase after starting {\it live operation}. The raw videos of the demonstration are available in the {\it online data repository}. 
In the demo, after robot operation begins, the operator decides to stop the robots' motion and request the addition of two new behaviors: shape deformation along the vertical axis (similar to the deformation shown in Fig.~\ref{fig:img1_samegestures}) and along the horizontal axis. The operator then observes that both of the new gestures are producing the same robot behavior, and requests a correction by using natural language to describe that the behaviors should be different. The next code generation step fixes the newly added SM, after which the operator's desired behaviors are performed successfully and the operator proceeds to complete the task. In the results (logs available in the {\it online data repository}), the first update step produces 1 new SM and adds 3 new nodes, with the full interaction and update taking 62\,s in total, and the second update modifies one line of code, taking 25\,s.

\subsection{Demo: Teleoperation of real robots}

\begin{figure}[htbp]
    \centering
    \begin{subfigure}[b]{0.32\textwidth}
        \centering
        \includegraphics[trim= 150 30 80 30, clip, width=\textwidth]{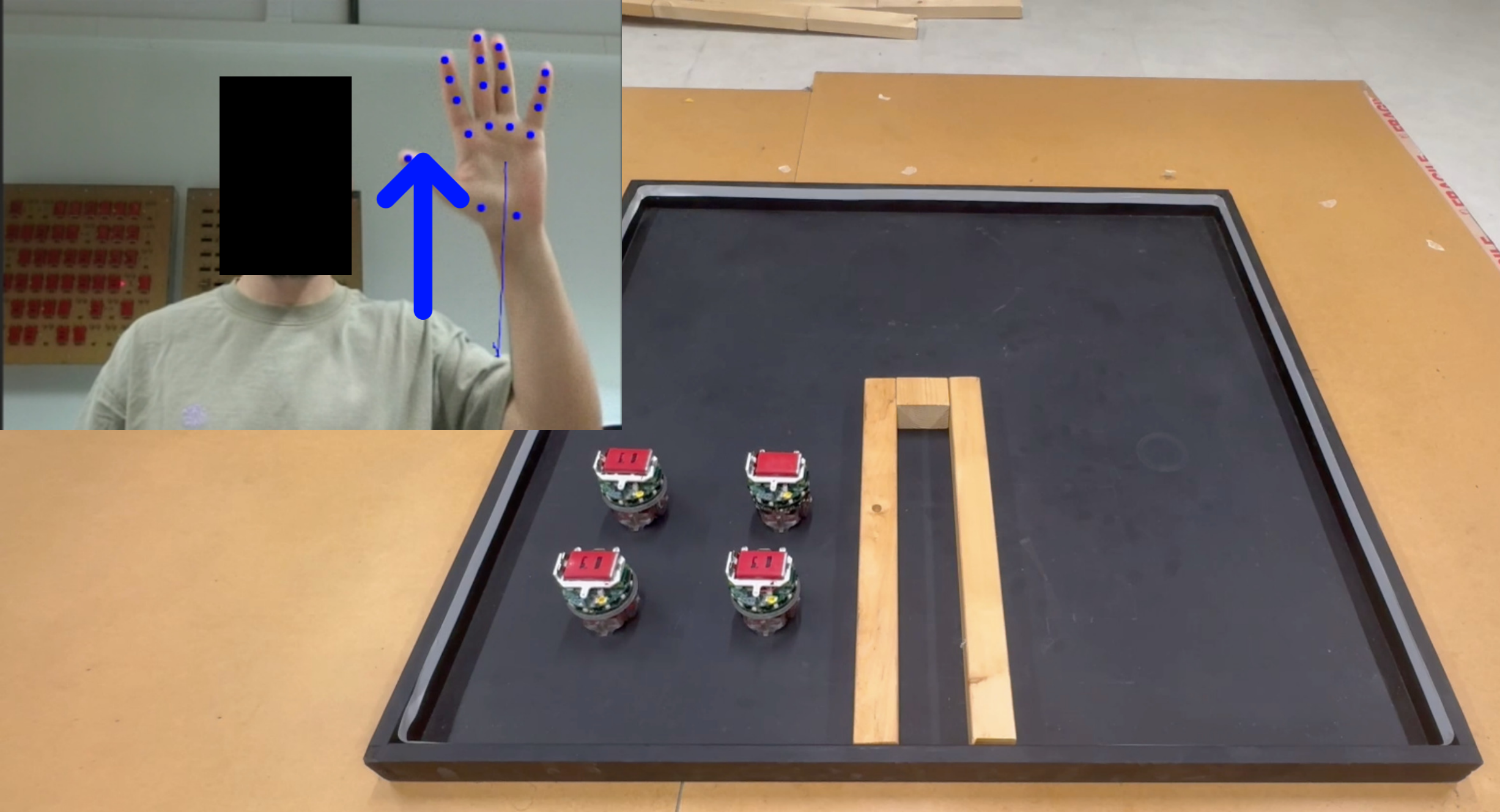}
        \caption{Starting navigation}
        \label{fig:img1_real_world_demo1}
        \vspace{1mm}
    \end{subfigure}
    \begin{subfigure}[b]{0.235\textwidth}
        \centering
        \includegraphics[trim= 150 30 80 30, clip, width=\textwidth]{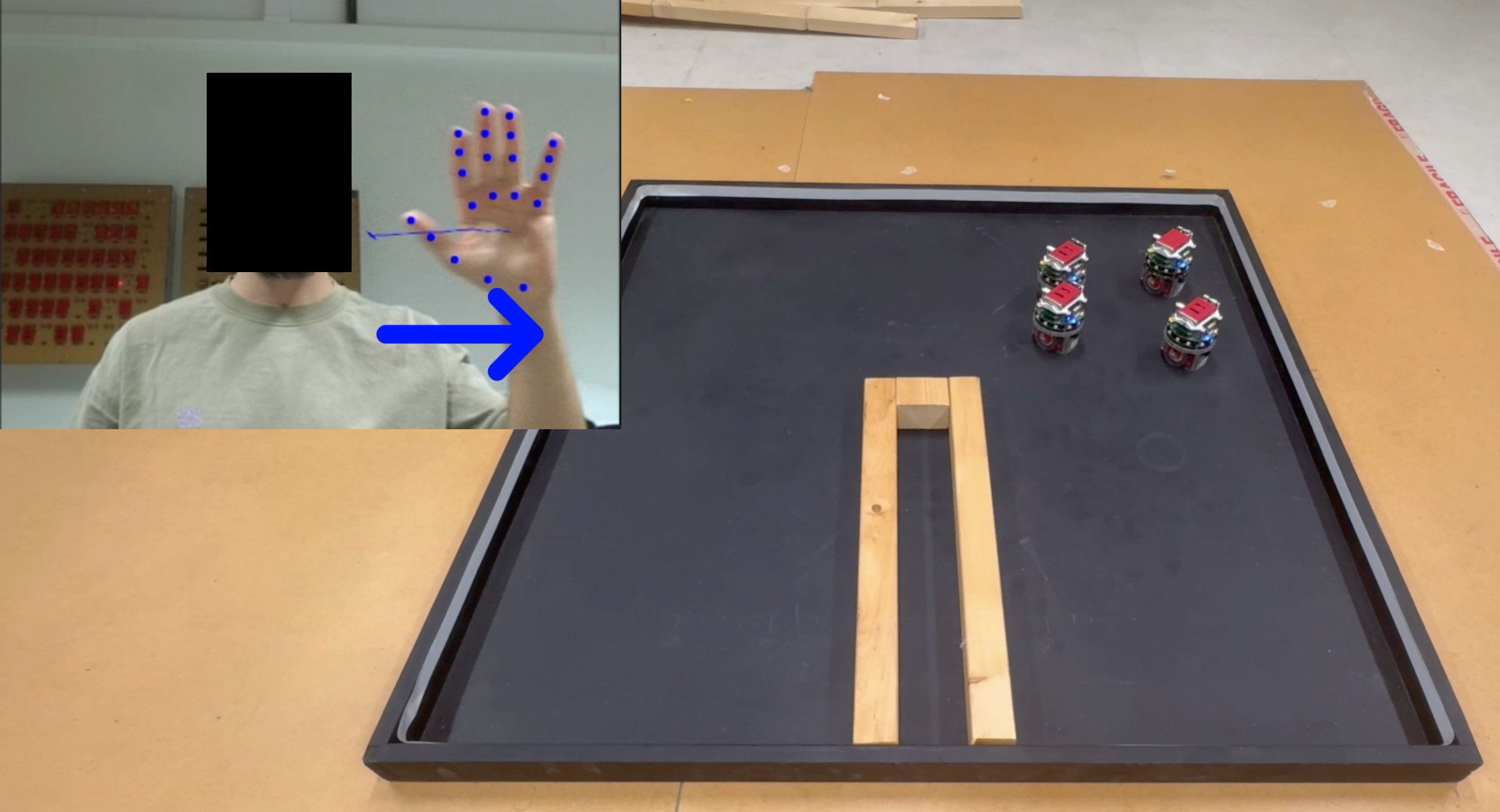}
        \caption{Navigating around obstacle}
        \label{fig:img2_real_world_demo1}
    \end{subfigure}
    \hfill
    \begin{subfigure}[b]{0.235\textwidth}
        \centering
        \includegraphics[trim= 150 30 80 30, clip, width=\textwidth]{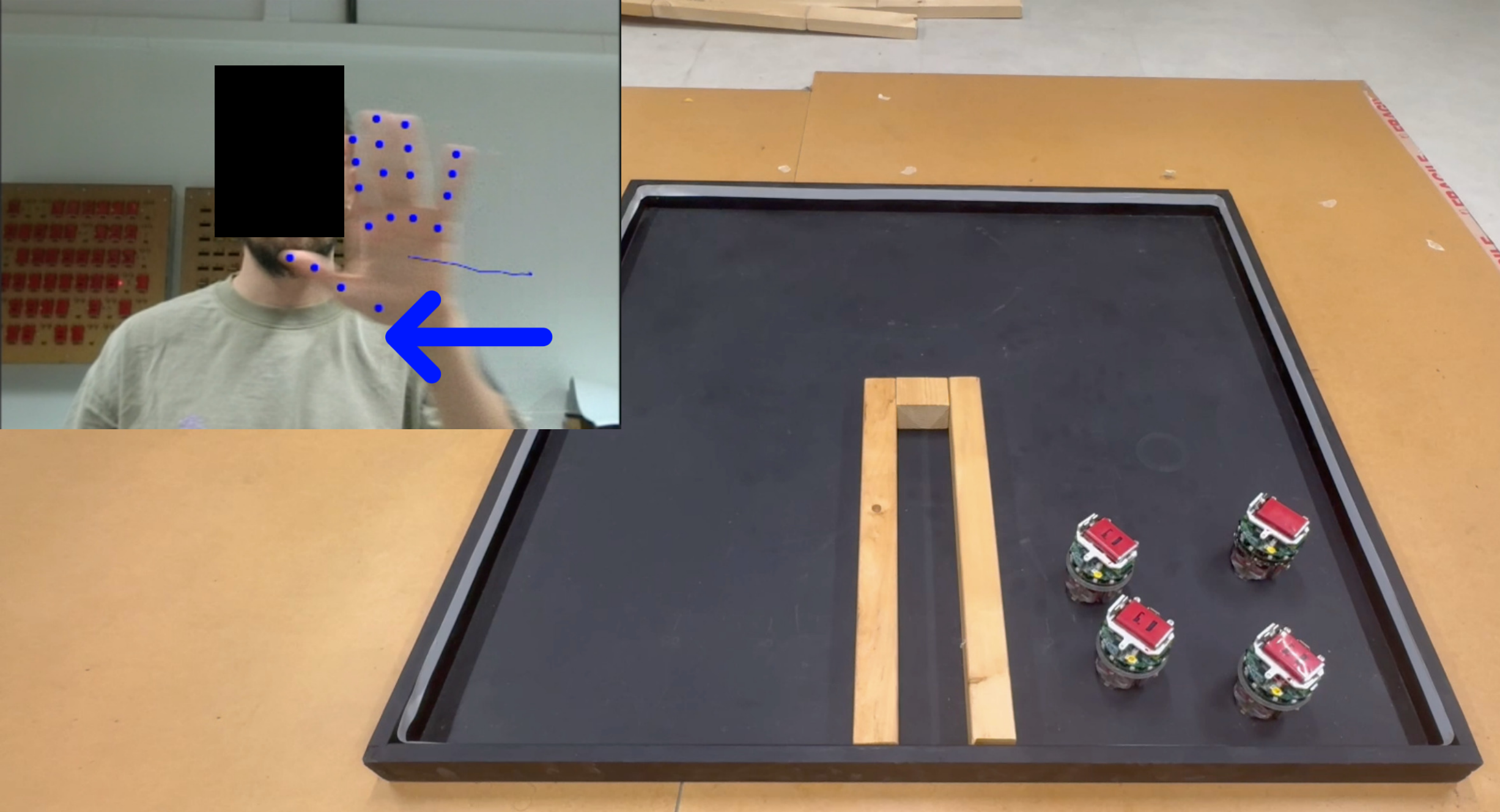}
        \caption{Reaching goal position}
        \label{fig:img3_real_world_demo1}
    \end{subfigure}
    \caption{{\bf Demo with real robots:} navigating an arena.}
    \label{fig:real_world_demo_1}
\end{figure}

\begin{figure}[htbp]
    \centering
    \begin{subfigure}[b]{0.235\textwidth}
        \centering
        \includegraphics[trim= 300 0 0 70, clip, width=\textwidth]{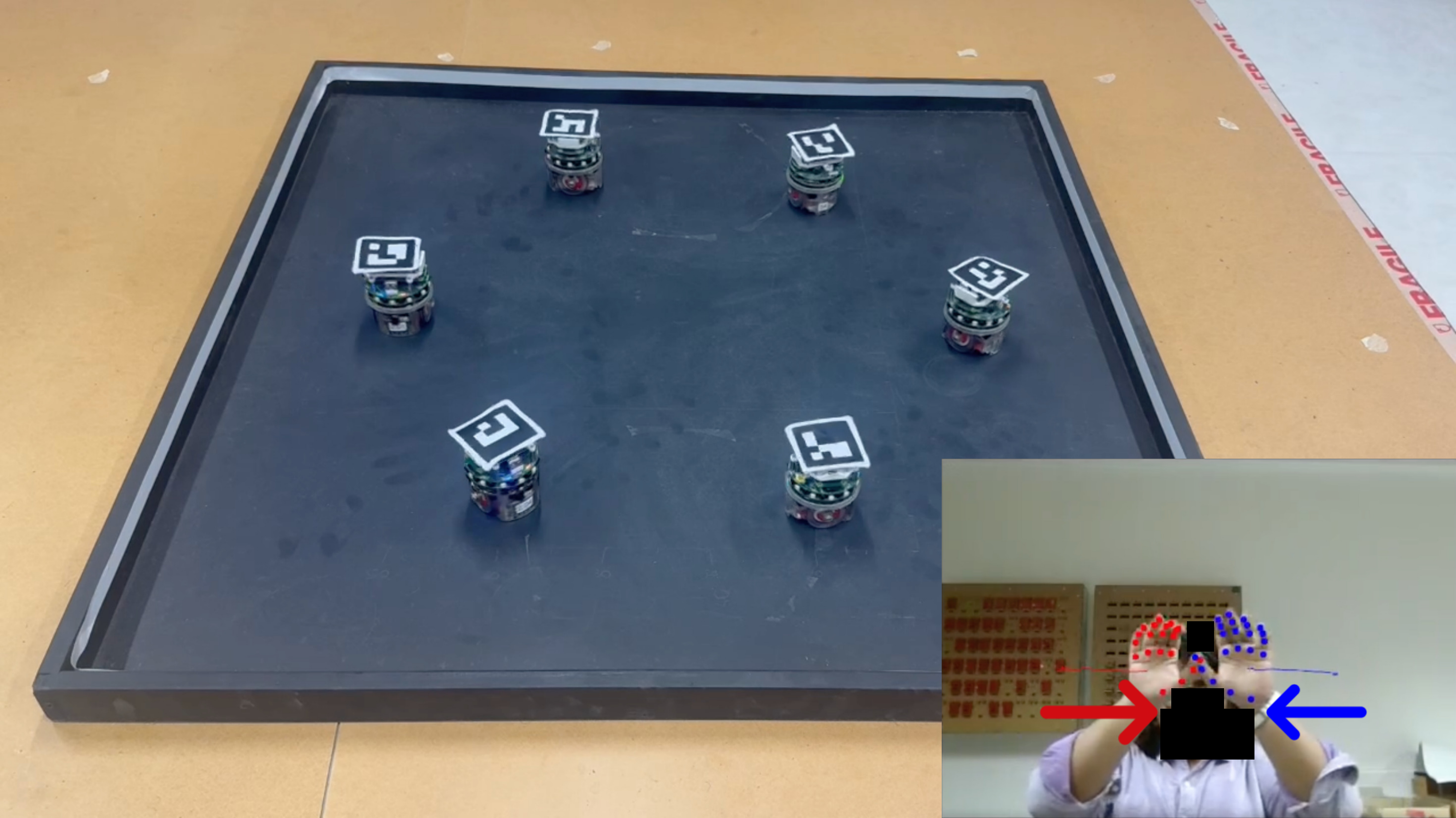}
        \caption{Starting shape: larger circle}
        \label{fig:img1_real_world_demo2}
    \end{subfigure}
    \hfill
    \begin{subfigure}[b]{0.235\textwidth}
        \centering
        \includegraphics[trim= 300 0 0 70, clip, width=\textwidth]{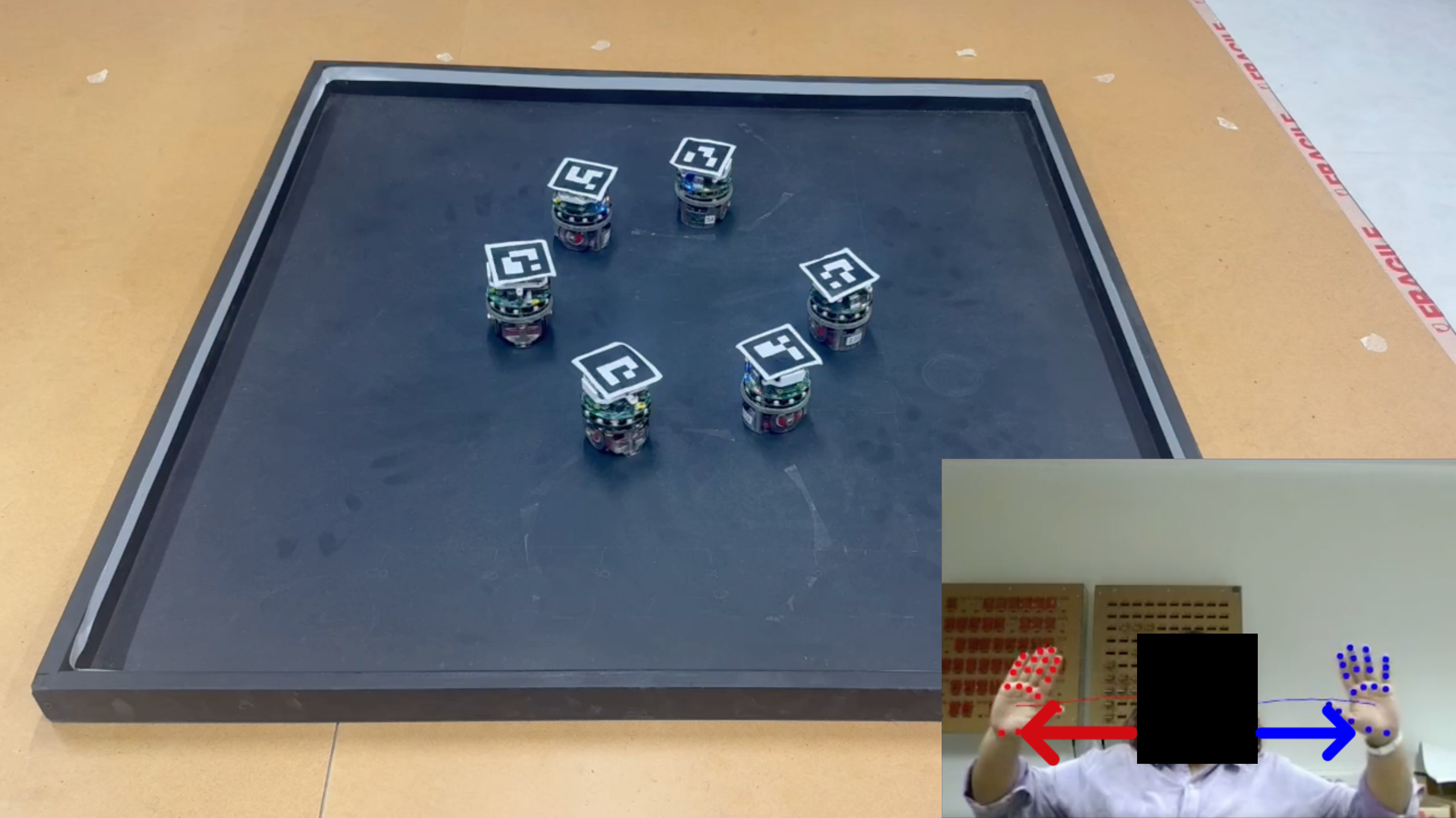}
        \caption{Ending shape: smaller circle}
        \label{fig:img2_real_world_demo2}
    \end{subfigure}
    \caption{{\bf Demo with real robots:} deforming a shape.}
    \label{fig:real_world_demo_2}
\end{figure}

We perform two demos with real e-puck~\cite{mondada2009puck} robots: one demo in which the operator forms 4 robots into an approximately square shape and navigates them through an arena (Fig.~\ref{fig:real_world_demo_1}) and one in which the operator forms 6 robots into an approximately circle shape and deforms the shape by changing its overall size (Fig.~\ref{fig:real_world_demo_2}). The operator interacts with the {\it Auto-HSI} prototype using a base station computer. On the base station computer, the nodes of the generated SMs send motor commands to the real robots wirelessly. 

\section{DISCUSSION AND FUTURE WORK}

A major current limitation is that robots are coordinated using external motion tracking and a central base station that issues motion commands. The most crucial future work is therefore to integrate with a swarm architecture that allows global targets to be disseminated from one robot to the rest of the swarm in a self-organized manner, while allowing the robots to maintain individual behaviors such as local collision avoidance. One architecture that delivers these capabilities is the {\it Self-organizing Nervous System} (SoNS) architecture~\cite{zhu2024self}; integration should be feasible, as the SoNS already provides an operator with a single access-point for programming behaviors, and communication from a SoNS-brain to an off-board LLM (without a human in the loop) has already been demonstrated~\cite{zhu2025online}. Future work could also investigate allowing a user to describe both individual robot and swarm-wide desired behaviors in natural language~\cite{nazzari2026tacos}.

Additionally, the current {\it Auto-HSI} runs on a base station computer, which then sends inputs to the robots. Therefore, another important aspect of future work is to transition to more capable robot platforms that have powerful onboard cameras, microphones, and processing, so {\it Auto-HSI} can run directly on the robots without any intermediary base station computer (but still assuming communication with an off-board LLM). Using such onboard equipment with a SoNS-enabled swarm, or a swarm with similar architecture capabilities, an operator would only need to stay in view and audio range of {\it any} robot in the swarm, rather than being restricted to a specific pre-allocated robot that is responsible for human interaction. 
Longer-term, if smaller-size LLMs and/or onboard processing capabilities improve sufficiently to support the required code generation based on natural language audio inputs using only onboard models, then the LLM can also run onboard the robots. 

Finally, transferring to real-world untrained operators and application scenarios with more open-ended gesture inputs and swarm behaviors would of course require further development: including of the gesture tracking, audio processing, emergency intervention mode, and expansion of the state machine generation mechanism to cover actuation beyond formation control. Furthermore, during real-world scenarios, operators are highly unlikely to have a bird-eye view of the robot swarm, as they did in this study. Therefore, the development of mechanisms providing feedback to the user about the swarm's state, performance, and environment would also be required.

%\section{CONCLUSIONS}

%A conclusion section is not required. Although a conclusion may review the main points of the paper, do not replicate the abstract as the conclusion. A conclusion might elaborate on the importance of the work or suggest applications and extensions. 

%%%%%%%%%%%%%%%%%%%%%%%%%%%%%%%%%%%%%%%%%%%%%%%%%%%%%%%%%%%%%%%%%%%%%%%%%%%%%%%%
\addtolength{\textheight}{-12cm}   % This command serves to balance the column lengths
%\section*{APPENDIX}

\section*{{\small ACKNOWLEDGMENT}}
{\small The authors would like to thank Giuseppe Patarino and Alexandre Pacheco for help with the real-robot demos.}
%Put sponsor acknowledgments in the unnumbered footnote on the first page.

%%%%%%%%%%%%%%%%%%%%%%%%%%%%%%%%%%%%%%%%%%%%%%%%%%%%%%%%%%%%%%%%%%%%%%%%%%%%%%%%

\bibliographystyle{ieeetr}
\bibliography{bibliography} 

\end{document}